\documentclass[10pt,twocolumn,letterpaper]{article}

\usepackage{iccv} 
\usepackage{multirow}
\usepackage{algorithm} 
\usepackage{algorithmic}
\usepackage{placeins}
\usepackage{afterpage}
\usepackage{cuted}

\definecolor{iccvblue}{rgb}{0.21,0.49,0.74}
\usepackage[pagebackref,breaklinks,colorlinks,allcolors=iccvblue]{hyperref}

\def\paperID{5785} 
\def\confName{ICCV}
\def\confYear{2025}

\title{Unified Geometry and Color Compression Framework for Point Clouds via Generative Diffusion Priors}

\author{Tianxin Huang \quad Gim Hee Lee\\
National University of Singapore\\
{\tt\small 21725129@zju.edu.cn, gimhee.lee@nus.edu.sg}}

\begin{document}
\maketitle
\begin{abstract}
With the growth of 3D applications and the rapid increase in sensor-collected 3D point cloud data, there is a rising demand for efficient compression algorithms. 
Most existing learning-based compression methods handle geometry and color attributes separately, treating them as distinct tasks, making these methods challenging to apply directly to point clouds with colors. Besides, the limited capacities of training datasets also limit their generalizability across points with different distributions.
In this work, we introduce a test-time unified geometry and color compression framework of 3D point clouds. Instead of training a compression model based on specific datasets, we adapt a pre-trained generative diffusion model to compress original colored point clouds into sparse sets, termed 'seeds', using prompt tuning. Decompression is then achieved through multiple denoising steps with separate sampling processes. 
Experiments on objects and indoor scenes demonstrate that our method has superior performances compared to existing baselines for the compression of geometry and color.
\end{abstract}    
\section{Introduction}
\label{sec:intro}
3D point clouds with color attributes are widely used in applications such as AR~\cite{suzuki2022augmented, singh2022augmented}, VR~\cite{garrido2021point, casado2023rendering}, and autonomous driving~\cite{reddy2018carfusion, geiger2013vision, cadena2016past}. With advancements in sensor and display technologies, the scale of 3D point clouds is rapidly increasing, placing greater demands on effective compression algorithms.
Since early works~\cite{huang20193d, wang2021lossy, wang2021multiscale}, learning-based methods have become a research focus, demonstrating strong performance in compressing the geometric structures of 3D point clouds. 
More recent studies~\cite{sheng2021deep, wang2022sparsea, mao2024pcac} have also explored the compression of attributes such as color, assuming the original coordinates are fully preserved, and designing networks to recover attributes from encoded features and coordinates.
However, most existing approaches still treat geometry~\cite{wang2022sparse, huang20223qnet} and color~\cite{sheng2021deep, 10.1145/3652212.3652217, wang2022sparsea} compression as separate tasks, handled through different frameworks. Geometry compression frameworks often disregard color information, while attribute compression frameworks assume uncompressed coordinates. Although some works~\cite{nguyen2023lossless, rudolph2024learned} have attempted to integrate geometry and attribute compression within a single framework, their generalizability remains limited to specific small-scale datasets. As a result, most existing approaches fail to provide a unified and generalizable compression method for both geometry structure and color attribute in 3D point clouds.
\begin{figure}[t]
	\begin{center}
		\scalebox{1.0}{
			\centerline{\includegraphics[width=\linewidth]{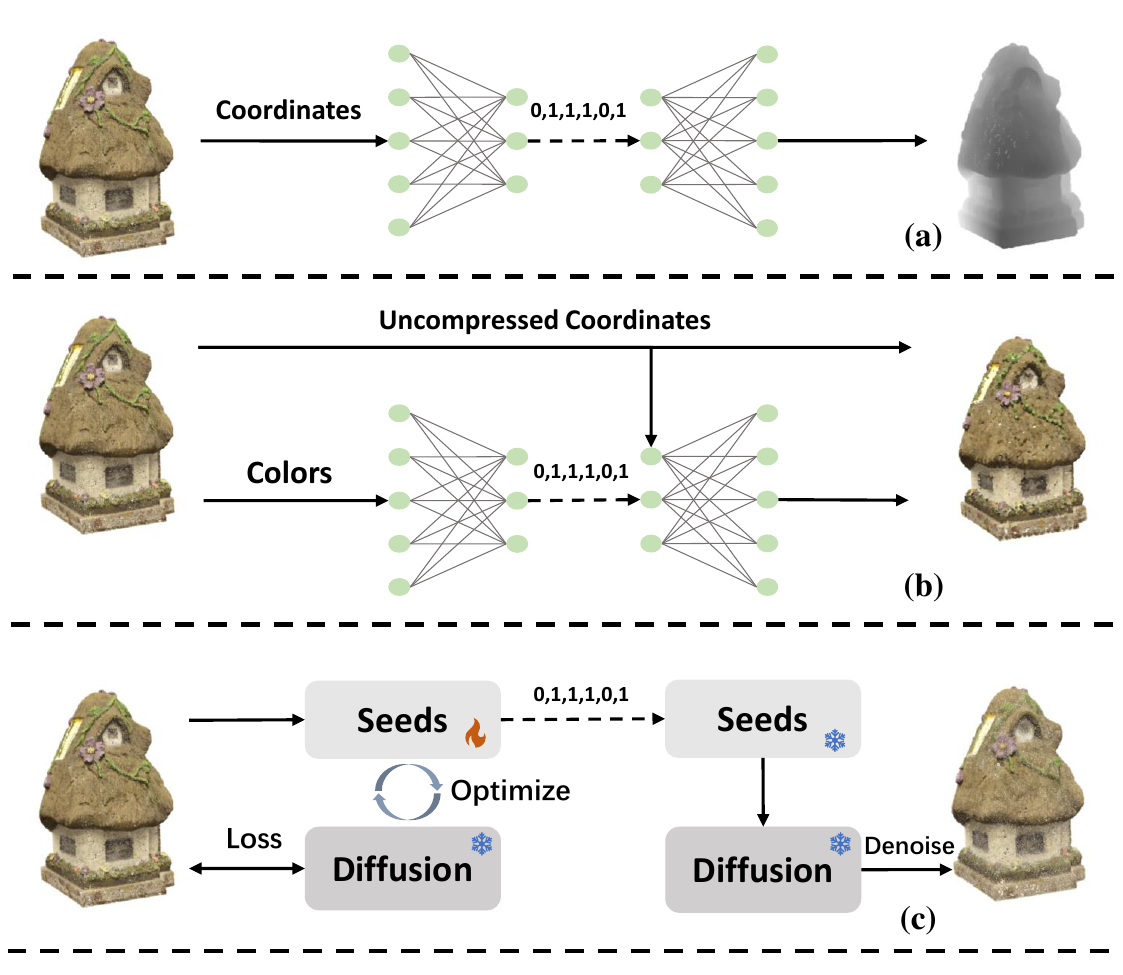}}
		}
		\vskip -0.1in
		\caption{Differences between our framework and existing point cloud compression methods. Conventional geometry compression techniques (a) focus solely on compressing coordinates, leaving color information unhandled. Attribute compression methods (b) encode colors into binary codes but retain uncompressed coordinates to assist with color decompression. In contrast, our framework (c) simultaneously optimizes the compression of both coordinates and colors into sparse sets, or '\textbf{seeds}.' These seeds are encoded into binary codes using a non-learning-based method and, after decoding, are used in a denoising process with a diffusion model for decompression.}
		\label{pic:intro}
	\end{center}
    \vspace{-0.2in}
\end{figure}

Point-E~\cite{nichol2022point} is a 3D generative model capable of generating 3D point clouds and colors from text descriptions or images. Benefited from training on its collected large amount of annotated point clouds, Point-E has great generalizability across point clouds with diverse categories and distributions. 
It comprises three diffusion models tailored to specific tasks: text-to-3D generation, image-to-3D generation, and point-to-point 3D up-sampling. The generative up-sampling model reconstructs high-resolution point clouds and colors from sparse inputs through a denoising process~\cite{ho2020denoising, nichol2022point}, leveraging rich prior knowledge to achieve coarse-to-fine transformation.

By learning continuous transformations from noise to structured points, this diffusion model can generate point clouds at various resolutions through the denoising process guided by sparse point sets, with multiple separate sampling processes, making it potentially applicable for point cloud compression by retaining only the sparse data needed to reconstruct dense point clouds.
Fine-tuning the pre-trained diffusion model of Point-E~\cite{nichol2022point} is a straightforward approach to apply it for compression, but it requires substantial time and computational resources to adapt the entire model to specific datasets.
To address this, we introduce prompt tuning for test-time optimization, adapting generative diffusion priors to specific point clouds for compression.
Prompt tuning~\cite{lester2021power,ruiz2023dreambooth} is used to optimize prompts for frozen diffusion models, aligning generated output more closely with the ground truth. In our case, sparse point sets act as prompts and would be optimized accordingly to improve the precision of decompressed dense point clouds.

In this work, we propose the first test-time unified compression framework for both geometry and colors of 3D point clouds. 
As shown in Figs.~\ref{pic:intro}-(a) and (b), most existing geometry and attribute compression methods handle coordinates and color attributes separately, treating them as distinct tasks.
As shown in Fig.~\ref{pic:intro}-(c), our method leverages the pre-trained up-sampling diffusion model from 3D generative model Point-E~\cite{nichol2022point} to simultaneously compress coordinates and colors, reducing the point clouds into sparse point sets, or 'seeds,' via prompt tuning. These seeds could be further encoded into binary codes and decoded using a non-learning-based method such as G-PCC~\cite{graziosi2020overview}.
Then, these seeds are decompressed into full-resolution 3D point clouds by denoising steps in multiple separate sampling processes.

This approach enables compression and decompression purely through the strong 3D priors of a pre-trained generative diffusion model, eliminating the need for additional dataset-specific training.
Prompt tuning during compression requires only short-term optimization, taking approximately 3 to 5 minutes on an RTX 4090 Ti for a 160k-point cloud. Additionally, we introduce a simple patch division strategy to better handle dense point clouds, effectively balancing compressed bit length and decompression accuracy.

Our contribution in this work can be summarized as:
\begin{itemize}
    \item We present a unified geometry and color compression framework based on a pre-trained diffusion model;
    \item By introducing prompt tuning for the pre-trained diffusion model, we propose the first test-time compression and decompression pipeline;
    \item Experiments on diverse data confirm that our method effectively handles 3D point clouds with colors, achieving superior performances compared to existing methods.
\end{itemize}

\section{Related Works}
\label{sec:related}

\subsection{3D Generation via Diffusion Models}

Following the success of 2D diffusion models in text-to-image generation~\cite{rombach2022high, saharia2022photorealistic, ho2020denoising}, 3D generation has gained significant research interest. To create 3D representations such as point clouds, meshes, NeRF~\cite{mildenhall2020nerf}, or 3D Gaussian primitives~\cite{kerbl20233d} from text or image inputs, researchers have explored two primary approaches.
The first approach involves adapting 2D priors for 3D generation~\cite{poole2022dreamfusion, wang2023score, mohammad2022clip, michel2022text2mesh, liu2023zero, tang2023dreamgaussian}. These methods typically optimize specific 3D representations using guidance from 2D diffusion models from different viewpoints, employing Score Distillation Sampling (SDS)~\cite{poole2022dreamfusion} through rendered images. These approaches leverage robust 2D pre-trained priors by fine-tuning view-guided diffusion models based on Stable Diffusion~\cite{rombach2022high}.
The second approach~\cite{jun2023shap, hu2024x, tochilkin2024triposr, nichol2022point} involves direct 3D generation by training on large-scale 3D datasets~\cite{chang2015shapenet, deitke2023objaverse}. These models offer improved efficiency and perform well with 3D objects, though scene-level generation still presents challenges due to limited data.
Among these methods, Point-E~\cite{nichol2022point} stands out for its ability to directly generate colored 3D point clouds from texts, images, or sparse point clouds, presenting promising potential for application in 3D point cloud compression.

\subsection{Point Cloud Compression}
3D point clouds consist of discrete points in 3D space without inherent topology. Existing point cloud compression methods can be broadly classified into non-learning-based and learning-based approaches.
Non-learning-based methods~\cite{cao20193d} typically use manually defined encoding rules and data structures, such as octrees~\cite{rusu20113d, graziosi2020overview} and kd-trees~\cite{galligan2018google}.
Learning-based geometry compression methods, on the other hand, seek to enhance compression by learning memory-efficient representations. Some approaches~\cite{wang2021lossy, wang2021multiscale, nguyen2021learning, quach2020improved} convert point clouds into voxel representations and use 3D CNNs to encode geometric structures into binary formats, though this may introduce precision loss. To mitigate this, point-based compression frameworks~\cite{He_2022_CVPR, huang20193d, huang20223qnet, wiesmann2021deep} directly operate on points, avoiding voxelization errors.

However, these learning-based methods primarily focus on geometry compression and lack the capability to process color attributes. In contrast, non-learning-based approaches such as Draco~\cite{galligan2018google}, G-PCC~\cite{graziosi2020overview}, and V-PCC~\cite{graziosi2020overview} support both geometry and color compression. Consequently, recent research~\cite{mao2024pcac, wang2022sparsea, sheng2021deep, 10.1145/3652212.3652217} has increasingly emphasized attribute compression for normals and colors.
Methods like SparsePCAC~\cite{wang2022sparsea} and ProgressivePCAC~\cite{10.1145/3652212.3652217} leverage sparse convolutions to encode attributes based on 3D coordinates, assuming the geometry remains uncompressed while compressing only colors or normals. Although some works~\cite{nguyen2023lossless, rudolph2024learned} attempt to compress both geometry and attributes simultaneously, they still require extensive training on specific datasets to adapt to different point cloud distributions, limiting their generalizability.

In this work, we propose a test-time unified compression framework for 3D point clouds with colors. Leveraging Point-E’s generative capabilities~\cite{nichol2022point}, our method decompresses both coordinates and colors through its denoising process while compressing the original point clouds into sparse sets via prompt tuning. We focus on the lossy compression~\cite{wang2021lossy, quach2020improved, mao2024pcac}, primarily evaluating our framework’s performance under constrained encoding bit budgets.
\begin{figure*}[t]
	\begin{center}
		\scalebox{1.0}{
			\centerline{\includegraphics[width=\linewidth]{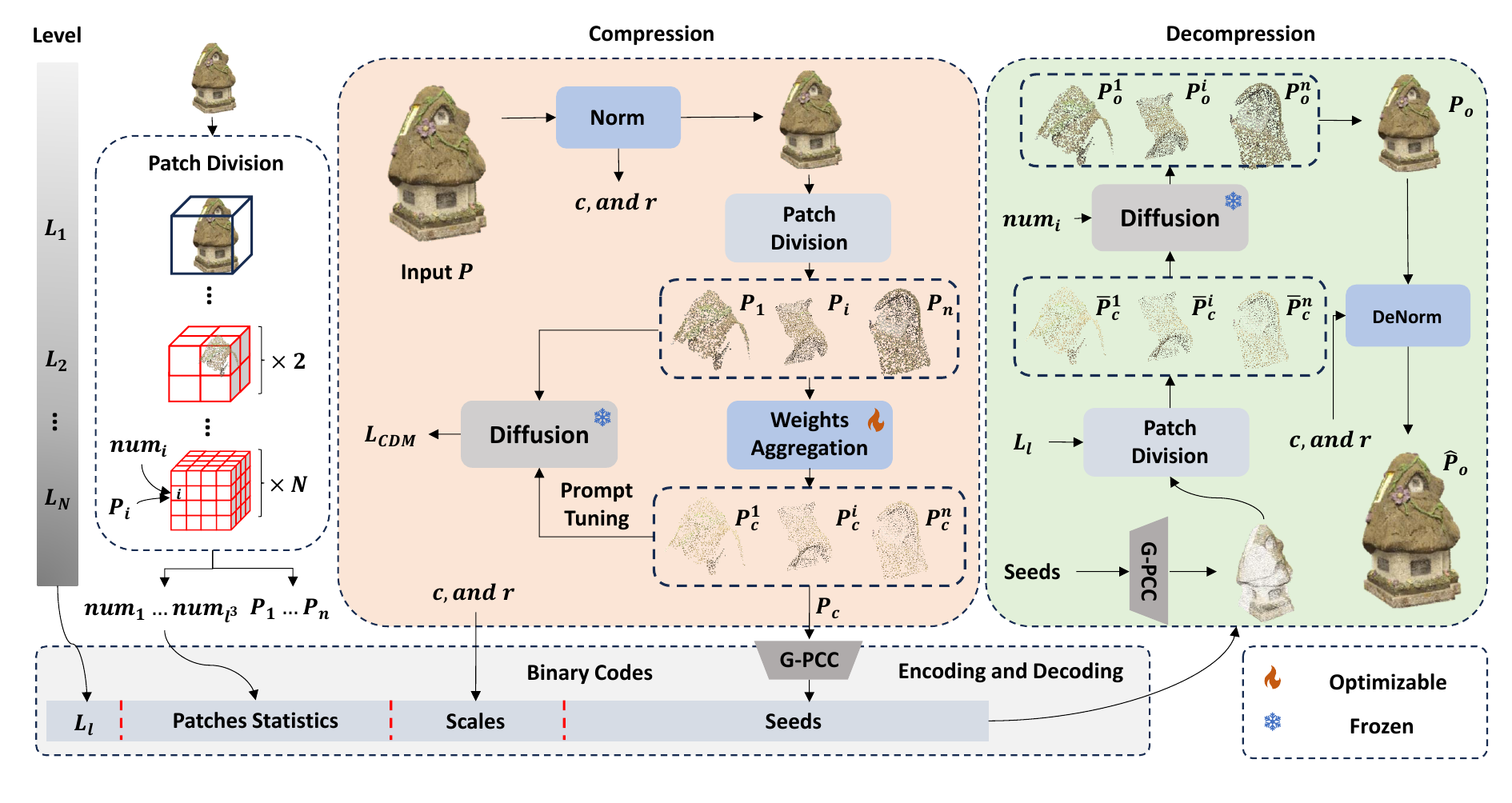}}
		}
		\vskip -0.2in
        \caption{The complete pipeline of our method. During compression, the input point cloud \( P \) is first normalized using the calculated center \( c \) and radius \( r \), then divided into \( n \) patches \( P_1, \dots, P_n \) according to the chosen compression level \( L_l \) using Patch Division. Seeds \( P_c^1, \dots, P_c^n \) for each patch are obtained through Weights Aggregation and optimized via prompt tuning using our proposed loss \( L_{CDM} \) computed on the pre-trained up-sampling diffusion model. Subsequently, \( L_l \), patch statistics (including patch resolutions), scales (center \( c \) and radius \( r \)), and seeds \( P_c = [P_c^1, \dots, P_c^n] \) are encoded into binary codes, which are then decoded for decompression. During decompression, the decoded seeds \( \bar{P}_c \) are split into patches \( \bar{P}_c^1, \dots, \bar{P}_c^n \) using patch division. Each patch \( \bar{P}_c^i \) is up-sampled to its recorded resolution \( num_i \) through the diffusion model’s denoising process, yielding \( P_o^i \). The concatenated output \( P_o = \{P_o^i | i = 1 \dots n\} \) is then denormalized to produce the final output \( \hat{P}_o \) using the recorded scales \( c \) and \( r \).}
		\label{pic:framework}
	\end{center}
	\vskip -0.2in
\end{figure*}
\section{Methodology}
\label{sec:method}
To accelerate the compression and decompression process, we only use first \( T \) steps from Point-E’s original diffusion-based up-sampling model~\cite{nichol2022point} in this work. As illustrated in Fig.~\ref{pic:framework}, our pipeline consists of Patch Division, Compression, Encoding and Decoding, and Decompression.

\subsection{Patch Division}


Point-E’s original diffusion-based upsampling model~\cite{nichol2022point} generates only sparse 3,072-point outputs from 1,024-point input prompts. To extend its capability to higher-resolution point clouds, we introduce a patch division strategy that segments the original point cloud into $N$ smaller patches.

As illustrated in Fig.~\ref{pic:framework}, the patch division is performed via simple voxelization. The input point cloud is partitioned into patches \(\{P_i | i=1, \dots, n\}\) using an \(l^3\) resolution grid at level \(L_l\). These patches are processed sequentially based on their xyz coordinates, and the number of points in each grid \(\{num_i | i=1, \dots, l^3\}\) is recorded to preserve resolution information for decompression. The segmented patches are then converted into seed points and encoded in the subsequent compression process. Although the number of grids appears to grow cubically, our experiments show that setting $l<=10$ is sufficient for most cases, ensuring that the encoded bits do not incur excessive memory overhead.

\subsection{Compression Process}
\noindent \textbf{Weights Aggregation.}
In this module, we first employ the boundary sampling (BDSam) method from \cite{huang20223qnet} to initialize the seeds for each \( i \)-th patch \( P_i \), ensuring that the boundary sizes of patches are preserved. During subsequent optimization, seeds positioned on the boundaries are detached to maintain the boundary integrity.
To mitigate the significant loss of local geometry and color information that can occur during sampling, we propose re-encoding each point in the initialized seeds \( P_c^s \) through weighted interpolation using \( k \) neighboring points from the original dense point cloud \( P \). This approach aggregates neighboring information to enhance the initialization quality. The \( k \) weights for the neighbor of each point in the seeds are initialized as an approximated one-hot array, where the weight for the original neighbor center is $1$ and others are $1e-4$.

Given the initialized sparse seeds \( P_{c}^s \), the KNN operation is used to obtain their \( k \) neighbors \( P_k = N_k(P_{c}^s, P) \) from the original point cloud \( P \), with weights \( W_c \in \mathbb{R}^{|P_c^s| \times k} \). 
The refined seeds \( P_c \) are then computed using a function \( g(\cdot) \) defined as follows:

\begin{equation}
\label{equ:phat}
    P_c = g(W_c, P_k) =  \sum \frac{|W_c|}{\sum |W_c|} \cdot P_k
\end{equation}

During subsequent prompt tuning, we further optimize \( W_c \) to more accurately capture the geometry and colors of the ground truth point cloud. By focusing on weight optimization rather than directly adjusting the seeds, we can maintain seed coordinates and colors within a reasonable range, promoting stability in the optimization process.

        
        
        
        
        

\begin{algorithm}[t]
	\caption{The Compression process.}
	\label{alg:com}
	\begin{algorithmic}[1]
        \STATE {Input: Divided patches $\hat{P}=[P_1, ..., P_n]$, the number of iterations $N_{iter}$, KNN operation $N_k(\cdot, \cdot)$.}
        \STATE Down-sample the patches with BDSam to initialized seeds $P_c$:
        \STATE Initialize $P_c=[], P_c^s=[]$
        \FOR{$i=1$ {\bfseries to} $n$}
         \STATE $P_c^s.insert(BDSam(P_i, 1024))$
         \ENDFOR
         \STATE $P_k = N_k(P_c^s, P)$, Initialize $W_c \in \mathbb{R}^{|P_c^s| \times k}$
         \FOR{$i=1$ {\bfseries to} $N_{iter}$}
         \STATE Randomly select $\text{idx} \sim \text{Uniform}(1, n)$
         \STATE Updating $W_c$ by descending gradient:
         
         $\nabla_{W_c} L_{CDM}(P_{idx},g(W_c, P_k)[idx], t)$
         \ENDFOR
         \STATE Calculate seeds $P_c = g(W_c, P_k)$
         \STATE {Output: the optimized seeds $P_c$}.

   \vskip -0.2in
	\end{algorithmic}
\end{algorithm}

\noindent \textbf{Prompt Tuning.}
After initialization, we apply prompt tuning as introduced by \cite{lester2021power, ruiz2023dreambooth} to refine the seeds. During prompt tuning, the parameters of Point-E’s diffusion-based model remain frozen, and only weights \( W_c \) are optimized through a loss calculated with the original point cloud \( P \).

Given the \( i \)-th patch out of \( n \) patches, \( P_i \), with corresponding seeds \( P_c^i \), and since the up-sampling model output is limited to 3072 points, we randomly sample 3072 points from \( P_i \) as the input \( x_0 \sim \text{RandomSample}(P_i, 3072) \) for the diffusion model.
Following \cite{rombach2022high, ho2020denoising}, the common diffusion model loss is defined as:
\begin{equation}
    \label{DMloss}
    L_{DM} = \mathbb{E}_{x_0,\epsilon \sim \mathcal{N}(0,1),t}[\|\epsilon - \epsilon_\theta(x_t, P_c^i , t)\|_2^2],
\end{equation}
where \( t \) is a time step uniformly sampled from \( \{1, 2, \dots, T\} \), and \( \epsilon_\theta(x_t, P_c^i, t) \) is the noise prediction network conditioned on the seeds \( P_c^i \) and time step \( t \). The noisy version \( x_t \) of the input \( x_0 \) at time step \( t \) is defined as:
\begin{equation}
\label{eq:xt}
x_t = \sqrt{\bar{\alpha}_t} x_0 + \sqrt{1 - \bar{\alpha}_t} \epsilon.
\end{equation}
Here, \( \alpha_t \) and $\bar{\alpha}_t$ denote constants selected as in \cite{ho2020denoising}.

As point clouds are permutation-invariant~\cite{qi2017pointnet}, altering the order of individual points does not affect the overall 3D shape. Chamfer Distance (CD)~\cite{fan2017point} is commonly used to measure shape differences between two point clouds through point-to-point matching.

From Eq.~\ref{DMloss}, we observe that the original diffusion model loss does not account for the permutation invariance of point clouds, implicitly assuming that the added noise \( \epsilon \) during noising and the predicted noise \( \epsilon_\theta(x_t, P_c^i, t) \) during denoising follow a consistent permutation. However, two similar point distributions do not necessarily share the same order.
To address this, we introduce a matching-based CD loss to relax this assumption. The Chamfer Distance is defined as:
\begin{equation}
\begin{split}
\mathcal{L}_{CD}(S_g,S_o)=\frac{1}{2}(\frac{1}{|S_g|}\sum_{x \in S_g}\mathop{\min}_{y \in S_o}\|x-y\|_2\\
+\frac{1}{|S_o|}\sum_{x \in S_o}\mathop{\min}_{y \in S_g}\|x-y\|_2),
\end{split}
\end{equation}
where \( S_g \) and \( S_o \) are two point clouds.

Given a time step \( t \), following Eq.~\ref{eq:xt}, the noised result at step \( t-1 \) is calculated as \( x_{t-1} = \sqrt{\bar{\alpha}_{t-1}} x_0 + \sqrt{1 - \bar{\alpha}_{t-1}} \epsilon \).
Our proposed Chamfer Diffusion Model Loss is defined as:
\begin{equation}
    L_{CDM}(P_i, t) = L_{CD}(x_{t-1}, \bar{x}_{t-1}),
\end{equation}
where \( \bar{x}_{t-1} \) represents the denoised distribution at step \( t-1 \), calculated as:
\begin{equation}
    \bar{x}_{t-1} = \frac{1}{\sqrt{\alpha_t}} \left( x_t - \frac{1 - \alpha_t}{\sqrt{1 - \bar{\alpha}_t}} \epsilon_\theta(x_t, P_c^i , t) \right).
\end{equation}
Here, \( L_{CDM} \) employs the Chamfer Distance loss to measure the difference between the noised distribution \( x_{t-1} \) and the denoised distribution \( \bar{x}_{t-1} \) at step \( t-1 \), mitigating potential issues caused by permutations.
As the $\bar{x}_0$ can also be estimated by $\bar{x}_0=\frac{x_t - \sigma_t \epsilon_\theta(x_t, t)}{\sqrt{\alpha_t}}$, we also try the loss:
\begin{equation}
    L_{inver} = L_{CD}(x_0, \bar{x}_0)
\end{equation}
Their performances are compared in Sec.~\ref{sec:aba_loss}.
After optimization, $W_c$ are transformed into seeds using Eq.~\ref{equ:phat} for encoding. The full compression pipeline is provided in Alg.~\ref{alg:com}.

\subsection{Encoding and Decoding}
The encoding and decoding process aims to achieve a lossless transformation between the decompression information and binary codes. As illustrated in Fig.~\ref{pic:framework}, the features to be encoded consist of four components: compression level $L_l$, patch statistics, scales, and seeds. The compression level and patch statistics are stored as a \( 1 \times 8 \text{ bits} \) char and \( l^3 \times 8 \text{ bits} \) chars, respectively. Scales, including the point cloud centers and radius, are saved as \( 4 \times 32 \text{ bits} \) floats. Seeds are concatenated, and compressed using G-PCC~\cite{graziosi2020overview} in a lossless manner.
In the decoding process, the compression level, patch statistics, scales, and seeds are sequentially decoded and prepared for usage in decompression.

\subsection{Decompression Process}
During decompression, the decoded seeds \( \bar{P}_c \) are divided into patches again using patch division, based on the decoded compression level \( L_l \). Each patch \( \bar{P}_c^i \) from the \( i \)-th of \( n \) patches is then used to generate the output point cloud \( P_o^i \) at a specific resolution \( num_i \), estimated from the patch statistics as shown in Fig.~\ref{pic:framework}.

The decompression \( f_D(\cdot) \) is carried out through a denoising process with multiple sampling steps, as described in Alg.~\ref{alg:f_D}. This process yields \( P_o^i = f_D(\bar{P}_c^i, num_i) \), and the full decompressed output is \( P_o = [P_o^1, \dots, P_o^n] \).
\begin{algorithm}[t]
	\caption{Decompression Process \( f_D (\bar{P}_c^i, num_i) \)}
	\label{alg:f_D}
	\begin{algorithmic}[1]
        \STATE {\textbf{Input:} Patch centers \( \bar{P}_c^i \), target resolution \( num_i \), constant \( \sigma_t \) (as defined in Point-E~\cite{nichol2022point}), Gaussian noise \( \epsilon \).}
        
        \STATE Calculate the number of iterations:
        \STATE \( N_s = \text{round}(num_i / 3072) \)
        \STATE Initialize \( P_o^i = [] \)
        
        \FOR{$i = 1$ {\bf to} $N_s$}
            \STATE Sample \( x_T = \sqrt{\bar{\alpha}_T} \bar{P}_c^i + \sqrt{1 - \bar{\alpha}_T} \epsilon \)
            \STATE Denoise the sampled $x_T$:
            \FOR{$t = T$ {\bf to} $1$}
                \STATE Update \( x_t = \frac{1}{\sqrt{\alpha_t}} \left( x_t - \frac{1 - \alpha_t}{\sqrt{1 - \bar{\alpha}_t}} \epsilon_\theta(x_t, \bar{P}_c^i, t) \right) \)
                \IF {$t > 1$}
                    \STATE Add noise \( x_t = x_t + \sigma_t \epsilon \)
                \ELSE
                    \STATE Set \( x_0 = x_t \)
                \ENDIF
            \ENDFOR
            \STATE Append \( x_0 \) to \( P_o^i \)
        \ENDFOR
        
        \STATE \textbf{Output:} Decompressed results \( P_o^i \)
	\end{algorithmic}
\end{algorithm}


Finally, the decompressed point cloud is de-normalized to its original scale using the decoded center \( c \) and radius \( r \) as follows:
\begin{equation}
    \hat{P}_o = [P_o^g \cdot r + c, P_o^c].
\end{equation}
\( P_o^g \) and \( P_o^c \) denotes coordinates and colors, respectively.

\section{Experiments}
\subsection{Dataset and Implementation Details}
To comprehensively evaluate our method, we conduct unified compression experiments on colored point clouds from point cloud object models~\cite{liu2022perceptual} and indoor scenes~\cite{dai2017scannet}, as well as geometry compression experiments~\cite{huang20223qnet}.
Unlike the voxelized 8iVFB dataset~\cite{d20178i}, we construct evaluation data by randomly sampling 160k colored points from objects and indoor scenes following 3QNet~\cite{huang20223qnet}, which we believe better represents real-world point cloud distributions.

Our implementation is based on PyTorch, with seeds optimized using the Adam optimizer at a learning rate of 1e-3 for approximately 1000 iterations during prompt tuning, without requiring re-training on specific datasets. Additional implementation details are available in the supplementary.
To validate our method’s effectiveness, we include comparisons with representative non-learning methods, Draco~\cite{galligan2018google} and G-PCC~\cite{graziosi2020overview}, as baselines. 
Although methods like CNet~\cite{nguyen2023lossless} claim to compress both geometry and color attributes, they do not provide open-source repositories or pretrained models.
Therefore, we construct learning-based baselines for unified compression by manually integrating geometry and attribute compression methods.
Specifically, we use widely used open-source geometry compression methods, PCGCv2~\cite{wang2021multiscale} and SparsePCGC~\cite{wang2022sparse} to compress coordinates, and the latest open-source attribute compression framework ProgressivePCAC~\cite{10.1145/3652212.3652217} to compress colors. The encoded bits from both components are then combined to form a unified representation.
\begin{figure*}[t]
	\begin{center}
		\scalebox{1.0}{
			\centerline{\includegraphics[width=\linewidth]{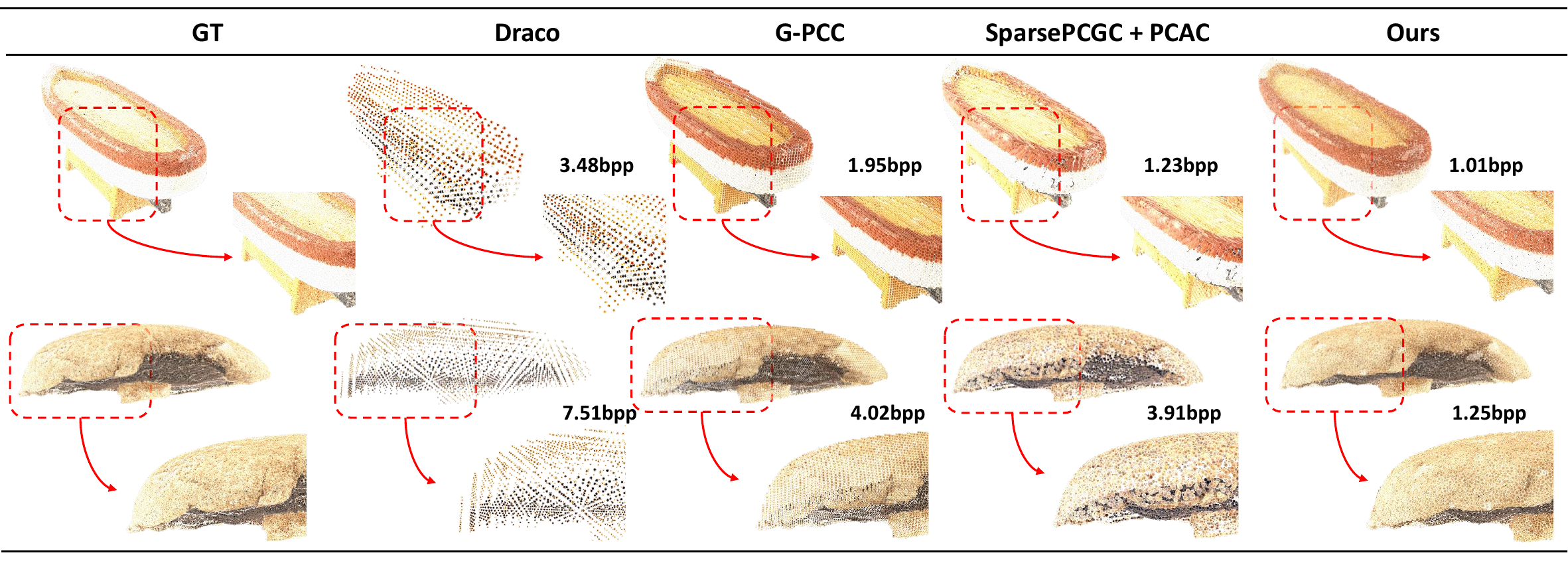}}
		}
		\vskip -0.1in
		\caption{Qualitative comparisons on 3D objects with colors.}
		\label{pic:objresults}
	\end{center}
	\vskip -0.3in
\end{figure*}
\subsection{Metrics}
Following prior work~\cite{quach2020improved, wang2021lossy, wang2021multiscale}, we evaluate compression bit rate using bits per point (bpp) and reconstruction quality using Peak Signal-to-Noise Ratio (PSNR). The bpp metric, defined as the average number of encoding bits per point, is calculated as:
\begin{equation}
\text{bpp} = \frac{L}{N},
\end{equation}
where \(L\) is the size of the compressed binary, and \(N\) is the number of points in the compressed point cloud.
For geometry evaluation, inspired by \cite{tian2017geometric, huang20223qnet}, PSNR is defined as:

\begin{equation}
\text{PSNR} = 10 \log_{10} \left(\frac{\max_{x \in S_1} \| x - \hat{x} \|_2}{Dis(S_1, S_2)}\right),
\end{equation}
where \(S_1\) and \(S_2\) represent the ground truth and decompressed point clouds, respectively, and \(\hat{x}\) is the nearest neighbor of \(x\) in \(S_1\). \(Dis(S_1, S_2)\) is the two-way average distance between \(S_1\) and \(S_2\), measured by either point-to-point or point-to-plane errors. We denote PSNR calculated from point-to-point or point-to-plane distances as \(PSNR_{D_1}\) and \(PSNR_{D_2}\), respectively.

For color evaluation, PSNR is defined as:

\begin{equation}
\text{PSNR} = 10 \log_{10} \left(\frac{255^2}{Dis(C_1, C_2)}\right),
\end{equation}
where \(C_1\) and \(C_2\) are the color attributes of \(S_1\) and \(S_2\). The color distance \(Dis(C_1, C_2)\) is calculated as the average color difference between each point and its nearest neighbor in YUV space. The PSNR for colors is noted as \(PSNR_{D_3}\).

Since PSNR varies with bpp as encoded information changes, we utilize BD-PSNR as a quantitative metric to assess our method’s relative improvement over others, following \cite{bjontegaard2001calculation, quach2020improved}.  BD-PSNR based on \(PSNR_{D_1}\) and \(PSNR_{D_2}\) reflects geometry performances, while that based on \(PSNR_{D_3}\) assesses color compression quality.


\subsection{Comparisons on Colored Objects}
In this section, we evaluate the performance of our method on 3D objects with color attributes. Quantitative results are shown in Table~\ref{tab:objs}, where performance is measured using BD-PSNR to capture relative improvements over other colored point cloud compression baselines. \textbf{Please note that our method does not have a direct entry in Table~\ref{tab:objs} since BD-PSNR represents relative improvements, with '+' and '-' indicating increases and decreases compared to existing methods, respectively}.

The results show significant improvements in geometry metrics PSNR-D1 and PSNR-D2 over existing baselines, while maintaining slightly improved performance on color metrics. 
Qualitative comparisons are shown in Fig.~\ref{pic:objresults}. Non-learning-based methods like Draco and G-PCC often exhibit regular distortions due to direct octree-based quantization and coordinate encoding, while the PCGCv2+PCAC baseline constructed for comparison displays artifacts such as holes. In contrast, our method produces smooth decompressed results with lower bpp.

\begin{table}[ht]
        \centering
        \normalsize
        \setlength\tabcolsep{1.3pt}
		\renewcommand\arraystretch{1.1}
  \scalebox{0.64}{
\begin{tabular}{c|ccccc}
\toprule
Objects                       &    & PCGCv2+PCAC     & G-PCC           & Draco     & SparsePCGC+PCAC      \\ \midrule
\multirow{3}{*}{ship}&D1&+12.22(51.12\%)&+11.35(44.41\%)&+32.84(61.03\%)&+4.61(24.79\%) \\ 
&D2&+11.52(42.15\%)&+10.98(43.42\%)&+30.51(59.71\%)&+1.33(5.95\%) \\ 
&D3&+9.08(25.43\%)&+3.02(10.08\%)&+15.06(47.98\%)&+3.08(10.51\%)\\ \midrule
\multirow{3}{*}{mushroom}&D1&+11.32(39.43\%)&+9.71(33.74\%)&+21.74(48.51\%)&+7.91(25.69\%) \\ 
&D2&+7.83(24.50\%)&+10.50(32.50\%)&+20.03(44.89\%)&+4.39(12.51\%) \\ 
&D3&+4.20(14.64\%)&+1.51(5.15\%)&+7.77(28.93\%)&+5.66(17.88\%)\\ \midrule
\multirow{3}{*}{statue}&D1&+10.70(59.91\%)&+7.81(39.91\%)&+21.12(47.80\%)&+5.46(29.24\%) \\ 
&D2&+6.80(32.26\%)&+8.06(39.53\%)&+18.98(43.69\%)&+1.67(7.28\%) \\ 
&D3&+5.44(18.20\%)&+0.25(0.89\%)&+7.42(28.68\%)&+3.56(11.91\%)\\ \midrule
\multirow{3}{*}{litchi}&D1&+7.11(51.77\%)&+8.63(40.24\%)&+26.33(53.02\%)&+5.54(31.48\%) \\ 
&D2&+3.48(22.62\%)&+7.62(36.50\%)&+22.70(49.14\%)&+2.41(11.38\%) \\ 
&D3&+7.28(20.94\%)&+3.02(9.30\%)&+10.70(32.52\%)&+10.86(27.85\%)\\ \midrule
\multirow{3}{*}{pencontainer}&D1&+15.32(44.24\%)&+8.14(27.16\%)&+24.90(54.82\%)&+11.83(32.18\%) \\ 
&D2&+12.76(33.01\%)&+8.32(24.91\%)&+23.38(52.55\%)&+9.35(22.18\%) \\ 
&D3&+8.50(28.86\%)&+1.05(4.02\%)&+10.41(38.50\%)&+11.19(34.36\%)\\ \midrule
\multirow{3}{*}{pineapple}&D1&+10.38(55.24\%)&+9.34(44.57\%)&+25.90(53.93\%)&+7.39(33.85\%) \\ 
&D2&+4.79(21.84\%)&+9.90(44.05\%)&+24.51(50.05\%)&+3.76(14.09\%) \\ 
&D3&+3.43(12.52\%)&+1.48(5.25\%)&+9.26(32.24\%)&+8.55(25.87\%)\\ \midrule
\multirow{3}{*}{pear}&D1&+13.25(65.58\%)&+10.61(47.05\%)&+27.39(54.37\%)&+7.33(45.74\%) \\ 
&D2&+7.42(35.06\%)&+13.65(51.33\%)&+28.44(53.00\%)&+2.18(9.96\%) \\ 
&D3&+8.35(23.18\%)&+2.28(6.22\%)&+10.51(28.44\%)&+14.30(36.98\%)\\ \midrule
\multirow{3}{*}{bumbameuboi}&D1&+10.70(38.23\%)&+5.99(23.24\%)&+20.80(47.81\%)&+8.22(29.27\%) \\ 
&D2&+7.52(22.90\%)&+7.45(25.02\%)&+19.88(45.43\%)&+1.44(4.61\%) \\ 
&D3&+2.60(10.03\%)&-1.21(4.97\%)&+5.26(21.69\%)&+3.00(11.34\%)\\ \midrule
\multirow{3}{*}{house}&D1&+8.25(69.17\%)&+8.32(40.74\%)&+26.83(55.54\%)&+3.88(28.11\%) \\ 
&D2&+4.12(28.52\%)&+8.15(39.14\%)&+24.64(51.81\%)&+0.34(1.94\%) \\ 
&D3&+3.37(12.68\%)&+1.36(5.02\%)&+10.94(38.96\%)&+4.40(14.76\%)\\ \midrule
\multirow{3}{*}{Aver}&D1&+11.03(52.74\%)&+8.88(37.90\%)&+25.32(52.98\%)&+6.91(31.15\%) \\ 
&D2&+7.36(29.20\%)&+9.40(37.38\%)&+23.67(50.03\%)&+2.99(9.99\%) \\ 
&D3&+5.81(18.50\%)&+1.42(4.55\%)&+9.70(33.11\%)&+7.18(21.27\%)\\ \bottomrule
\end{tabular}
}
\caption{Quantitative comparisons on Colored Objects.  \textbf{Items in the table denote the relative improvements of our method over other methods measured with BD-PSNR. '+' and '-' indicates performance increases and decreases, respectively.}}
\vskip -0.2in
\label{tab:objs}
\end{table}

\subsection{Comparisons on Indoor Scenes}
\begin{figure*}[t]
	\begin{center}
		\scalebox{1.0}{
			\centerline{\includegraphics[width=\linewidth]{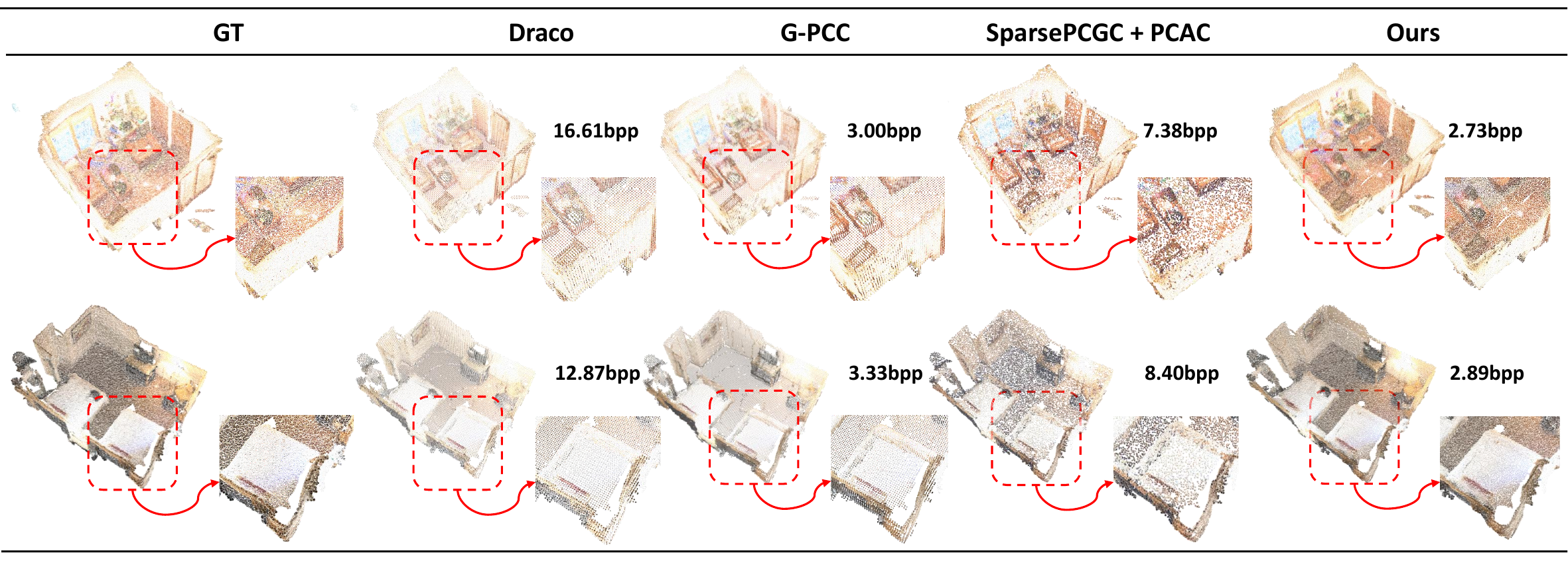}}
		}
		\vskip -0.1in
		\caption{Qualitative comparisons on 3D indoor scenes with colors.}
		\label{pic:scannet}
	\end{center}
	\vskip -0.15in
\end{figure*}

To further assess our method’s performance on real scanned point cloud data, we evaluate it on indoor scenes from ScanNet~\cite{dai2017scannet}. Qualitative and quantitative comparisons are shown in Fig.~\ref{pic:scannet} and Table~\ref{tab:indoors}, respectively. 
As illistrated in Fig.~\ref{pic:scannet}, our method produces smoother results closer to the ground truth with lower bpp. The results in Table~\ref{tab:indoors} demonstrate that our method still achieves obvious improvements over existing baseline models, revealing its generalizability from object-level to scene-level point clouds.

\begin{table}[th]
        \centering
        \normalsize
        \setlength\tabcolsep{1.5pt}
		\renewcommand\arraystretch{1.1}
  \scalebox{0.7}{
\begin{tabular}{c|ccccc}
\toprule
Objects                       &    & PCGCv2+PCAC     & G-PCC           & Draco     & SparsePCGC+PCAC      \\ \midrule
\multirow{3}{*}{scene0}&D1&+8.60(57.69\%)&+7.84(34.92\%)&+24.52(56.19\%)&+0.33(2.19\%) \\ 
&D2&+4.82(26.18\%)&+6.68(31.64\%)&+22.71(70.73\%)&-2.58(13.72\%) \\ 
&D3&+2.12(7.94\%)&+2.23(8.06\%)&+10.14(39.66\%)&-0.89(3.25\%)\\ \midrule
\multirow{3}{*}{scene1}&D1&+7.76(36.16\%)&+6.03(29.28\%)&+24.15(54.79\%)&+3.61(15.54\%) \\ 
&D2&+5.32(21.68\%)&+5.27(22.45\%)&+21.81(55.72\%)&+1.06(3.96\%) \\ 
&D3&+4.66(16.42\%)&+1.58(5.51\%)&+10.65(37.19\%)&+4.20(13.76\%)\\ \midrule
\multirow{3}{*}{scene2}&D1&+11.71(75.77\%)&+7.41(36.35\%)&+23.55(53.11\%)&+4.72(28.63\%) \\ 
&D2&+7.87(43.19\%)&+6.67(33.99\%)&+20.91(53.13\%)&+0.09(0.49\%) \\ 
&D3&+5.67(17.43\%)&+2.46(7.52\%)&+12.74(40.19\%)&+3.01(9.02\%)\\ \midrule
\multirow{3}{*}{scene3}&D1&+9.84(62.57\%)&+7.45(34.88\%)&+23.37(53.36\%)&+4.16(25.96\%) \\ 
&D2&+6.07(32.28\%)&+6.10(29.97\%)&+21.58(56.08\%)&+0.43(2.23\%) \\ 
&D3&+3.99(14.53\%)&+2.08(7.43\%)&+9.91(37.98\%)&+3.00(10.57\%)\\ \midrule
\multirow{3}{*}{scene4}&D1&+8.20(65.09\%)&+5.50(30.00\%)&+23.56(54.38\%)&+1.00(7.58\%) \\ 
&D2&+3.82(24.89\%)&+4.68(26.90\%)&+20.89(53.19\%)&-5.17(31.60\%) \\ 
&D3&+3.20(11.34\%)&+0.11(0.39\%)&+11.26(40.17\%)&-2.38(7.98\%)\\ \midrule
\multirow{3}{*}{scene5}&D1&+10.88(71.09\%)&+8.90(39.64\%)&+27.36(56.41\%)&+3.54(24.70\%) \\ 
&D2&+6.49(34.38\%)&+8.61(39.45\%)&+24.29(56.74\%)&-1.06(5.95\%) \\ 
&D3&+8.66(26.39\%)&+3.49(10.94\%)&+13.93(43.82\%)&+3.42(10.81\%)\\ \midrule
\multirow{3}{*}{Aver}&D1&+9.50(61.40\%)&+7.19(34.18\%)&+24.42(54.71\%)&+2.89(17.43\%) \\ 
&D2&+5.73(30.44\%)&+6.33(30.73\%)&+22.03(57.60\%)&-1.20(7.43\%) \\ 
&D3&+4.72(15.67\%)&+1.99(6.64\%)&+11.44(39.83\%)&+1.73(5.49\%)\\ \bottomrule
\end{tabular}
}
\vskip -0.05in
\caption{Quantitative comparisons on indoor scenes.  Items in the table denote the relative improvements of our method over other methods measured with BD-PSNR. '+' and '-' indicates performance increases and decreases, respectively.}
\vskip -0.15in
\label{tab:indoors}
\end{table}

\subsection{Comparison on Geometry Compression}
\label{sec:geocom}
Although our method is designed to simultaneously compress the geometry and colors of point clouds, it can also be applied to pure geometry compression by assigning constant colors, such as white or black, to the input point clouds.
In this section, we conduct a simple comparison between our method and several representative geometry compression approaches by setting the point cloud colors to white. The quantitative results are shown in Table~\ref{tab:geo}. 

The results demonstrate that our method outperforms existing geometry compression baselines while requiring no dataset-specific training, unlike methods such as 3QNet~\cite{huang20223qnet} and PCGCv2~\cite{wang2021multiscale}. 
These improvements can be attributed to the strong generalization capability of the diffusion-based up-sampling model in Point-E~\cite{nichol2022point}, which reconstructs continuous surfaces from optimized sparse points, or 'seeds.' By encoding only the seeds for compression, our method eliminates the need for additional encoded features, as required by methods like 3QNet~\cite{huang20223qnet}, resulting in shorter compressed binary codes.

\begin{table}[ht]
        \centering
        \normalsize
        \setlength\tabcolsep{1.3pt}
		\renewcommand\arraystretch{1.0}
  \scalebox{0.75}{
\begin{tabular}{c|ccccc}
\toprule
Objects                       &  & 3QNet  & PCGCv2              & G-PCC & SparsePCGC          \\ \midrule
\multirow{3}{*}{eros}&D1&+5.44(31.74\%)&+25.43(69.36\%)&+5.18(27.34\%)&+9.26(43.57\%) \\ 
&D2&+1.87(8.09\%)&+13.93(59.85\%)&+4.17(16.01\%)&+5.87(22.52\%)\\ \midrule
\multirow{3}{*}{kitten}&D1&+7.80(38.40\%)&+37.48(76.12\%)&+3.90(21.40\%)&+6.51(33.67\%) \\ 
&D2&+8.67(31.40\%)&+29.01(74.08\%)&+5.64(19.85\%)&+1.72(5.86\%)\\ \midrule
\multirow{3}{*}{boy01}&D1&+2.15(14.98\%)&+20.03(63.13\%)&+2.30(16.52\%)&+8.34(50.15\%) \\ 
&D2&+2.18(10.09\%)&+10.22(51.89\%)&+2.32(10.88\%)&+10.05(38.71\%)\\ \midrule
\multirow{3}{*}{julius}&D1&+0.64(4.29\%)&+21.08(60.57\%)&+5.43(32.15\%)&+6.56(34.86\%) \\ 
&D2&-2.12(10.00\%)&+7.61(40.55\%)&+5.46(22.87\%)&+0.41(1.58\%)\\ \midrule
\multirow{3}{*}{pig}&D1&+3.80(23.07\%)&+24.91(71.26\%)&+4.84(29.04\%)&+9.18(49.36\%) \\ 
&D2&+5.42(21.24\%)&+15.99(66.24\%)&+7.29(27.64\%)&+7.95(28.62\%)\\ \midrule
\multirow{3}{*}{biggirl}&D1&+2.62(16.77\%)&+21.34(67.34\%)&+2.18(13.98\%)&+8.71(52.02\%) \\ 
&D2&+3.50(15.73\%)&+14.09(54.67\%)&+2.58(10.85\%)&+10.20(39.42\%)\\ \midrule
\multirow{3}{*}{pierrot}&D1&+0.04(0.28\%)&+8.25(37.97\%)&+4.20(27.76\%)&+7.34(37.82\%) \\ 
&D2&+0.43(1.76\%)&-0.42(2.49\%)&+4.45(22.45\%)&+1.43(4.89\%)\\ \midrule
\multirow{3}{*}{nicolo}&D1&+0.42(2.90\%)&+19.50(57.31\%)&+1.45(9.72\%)&+4.17(23.81\%) \\ 
&D2&-1.90(9.42\%)&+7.14(34.26\%)&+0.16(0.75\%)&-4.04(15.85\%)\\ \midrule
\multirow{3}{*}{duck}&D1&-1.22(7.86\%)&+8.62(34.96\%)&+4.09(27.29\%)&+7.53(38.86\%) \\ 
&D2&-0.79(3.28\%)&+1.28(6.51\%)&+4.77(23.26\%)&+1.40(4.84\%)\\ \midrule
\multirow{3}{*}{star}&D1&+3.64(21.85\%)&+23.59(60.82\%)&+5.50(30.13\%)&+6.04(31.66\%) \\ 
&D2&+3.61(13.78\%)&+15.78(60.60\%)&+9.18(31.51\%)&+2.83(9.38\%)\\ \midrule
\multirow{3}{*}{Aver}&D1&+2.53(14.64\%)&+21.02(59.88\%)&+3.91(23.53\%)&+7.36(39.58\%) \\ 
&D2&+2.09(7.94\%)&+11.46(44.61\%)&+4.60(18.61\%)&+3.78(14.00\%)\\ \bottomrule
\end{tabular}
}
\caption{Quantitative comparisons on geometry compression. Items denote the relative improvements of our method over other methods measured with BD-PSNR. '+' and '-' indicates performance increases and decreases, respectively.}
\label{tab:geo}
\end{table}

\subsection{Ablation Study}
\textbf{Ablation study for different components}
To assess the impact of our proposed components, we conduct an ablation study in Table~\ref{tab:aba_com}. The terms *Decom*, *Weights*, and *Tuning* refer to the decompress, weights aggregation, and prompt tuning described in Sec.~\ref{sec:method}, respectively. *Seeds* denotes using seeds directly as the results without decompression.
Performance is evaluated using BD-PSNR of our method against G-PCC, calculated with \( PSNR_{D1} \), \( PSNR_{D2} \), and \( PSNR_{D3} \).
The results indicate that removing any component decreases overall performance, confirming that each component contributes to the final outcome.

\begin{table}[ht]
        \centering
        \normalsize
        \setlength\tabcolsep{1.5pt}
  \scalebox{1.0}{
\begin{tabular}{cccc|ccc}
\toprule
Seeds                     & Decom                     & Weights                   & Tuning                    & D1             & D2              & D3             \\ \midrule
\checkmark &                           &                           & \textbf{}                 & -0.99          & +5.49           & -2.97          \\
\checkmark & \checkmark &                           & \textbf{}                 & +8.27          & +8.67           & +0.91          \\
\checkmark & \checkmark & \checkmark & \textbf{}                 & +8.37          & +8.67           & +0.97          \\
\checkmark & \checkmark & \checkmark & \checkmark & \textbf{+8.88} & \textbf{+9.40} & \textbf{+1.42} \\ \bottomrule
\end{tabular}
}
\caption{Ablation Study for proposed components. Performances are evaluated with BD-PSNR of our method against G-PCC.}
\vskip -0.1in
\label{tab:aba_com}
\end{table}

\noindent\textbf{Ablation study for tuning losses}
\label{sec:aba_loss}
In this work, we introduce the Chamfer Diffusion Model Loss \( L_{CDM} \) to replace the original MSE-based diffusion model loss \( L_{DM} \). To assess its effectiveness, we compare various losses for prompt tuning, as shown in Table~\ref{tab:aba_loss}. The results, measured using BD-PSNR, evaluate our method against G-PCC. Here, \( L_{inver} \), \( L_{DM} \), and \( L_{CDM} \) refer to the inverse loss, diffusion model loss, and Chamfer diffusion model loss, respectively, as described in Sec.~\ref{sec:method}.
As shown in Table~\ref{tab:aba_loss}, our proposed \( L_{CDM} \) achieves the best performance in prompt tuning, while \( L_{inver} \) and \( L_{DM} \) even perform worse than the framework without prompt tuning on certain metrics. 
This confirms the permutation-invariance of point clouds limits \( L_{DM} \)'s effectiveness for point generation, whereas our proposed loss effectively addresses this limitation.
\begin{table}[ht]
        \centering
        \normalsize
        \setlength\tabcolsep{1.5pt}
  \scalebox{1.0}{
\begin{tabular}{c|cccc}
\toprule
Metrics & No tuning & $L_{inver}$ & $L_{DM}$    & $L_{CDM}$(ours)            \\ \midrule
D1      & +5.90     & +8.42   & +6.19  & \textbf{+8.88}  \\
D2      & +5.92     & +8.70  & +6.36 & \textbf{+9.40} \\
D3      & -2.81     & +0.62   & -2.49  & \textbf{+1.42}  \\ \bottomrule
\end{tabular}
}
\caption{Ablation Study for losses of prompt tuning. Performances are evaluated with BD-PSNR of our method against G-PCC.}
\vskip -0.2in
\label{tab:aba_loss}
\end{table}

\subsection{Discussion about the Efficiency}
\label{aba_eff}
In this section, we present a brief comparison of the efficiency of our method with other baselines, as shown in Table~\ref{tab:aba_eff}. 
While our method requires more time for compression due to prompt tuning, the preprocessing and decompression times remain affordable, and our approach requires no training on specific datasets while offering strong robustness. These characteristics make our method well-suited for offline applications such as cultural heritage digitization. Additionally, the efficiency of test-time prompt tuning could be further enhanced using techniques like Mixed Precision Training~\cite{micikevicius2017mixed} or tensor/model parallelism~\cite{rajbhandari2020zero}, potentially reducing its computational overhead.
\begin{table}[ht]
        \centering
        \normalsize
        \setlength\tabcolsep{1.5pt}
		\renewcommand\arraystretch{1.1}
  \scalebox{0.75}{
\begin{tabular}{cc|cccc}
\hline
\multicolumn{2}{c|}{Methods}                & Pre-process & Compress & Decompress & Training      \\ \hline
\multirow{2}{*}{Non-Learning} & Draco       & 0           & 0.30s    & 0.10s      & 0             \\
                              & G-PCC       & 0           & 0.50s    & 0.06s      & 0             \\ \hline
\multirow{2}{*}{Learning}     & PCGCv2+PCAC & 3.36s       & 1.55s    & 3.37s      &  days \\
                              & Ours        & 0.76s       & 4 min    & 6.05s      & 0             \\ \hline
\end{tabular}
}
\caption{Efficiency Comparison.}
\label{tab:aba_eff}
\vskip -0.15in
\end{table}

\section{Limitation}
As discussed in Sec.~\ref{aba_eff}, while our method eliminates the need for training, prompt tuning results in a longer compression time compared to existing learning-based approaches. However, it remains well-suited for offline compression scenarios where real-time processing is not required, while its efficiency may also be further improved through parallelism or mix-precision techniques. We will explore it in our future work.
Moreover, to the best of our knowledge, our method is the first test-time framework for unified geometry and color compression in point clouds, offering significant potential for further optimization and exploration.

\section{Conclusion}
In this work, we proposed a unified test-time compression framework for both geometry and color in 3D point clouds, leveraging priors from a pre-trained generative diffusion model. Unlike conventional methods that treat geometry and color compression as separate tasks, our approach integrates both using Point-E~\cite{nichol2022point}’s upsampling model. By generating sparse 'seeds' through prompt tuning, our method enables high-fidelity decompression without dataset-specific training. Experimental results demonstrate that our framework achieves superior compression performance across diverse datasets.

{
    \small
    \bibliographystyle{ieeenat_fullname}
    \bibliography{main}

\begin{thebibliography}{56}
\providecommand{\natexlab}[1]{#1}
\providecommand{\url}[1]{\texttt{#1}}
\expandafter\ifx\csname urlstyle\endcsname\relax
  \providecommand{\doi}[1]{doi: #1}\else
  \providecommand{\doi}{doi: \begingroup \urlstyle{rm}\Url}\fi

\bibitem[bui()]{buildings}
Gti-upm point-cloud data set.
\newblock \url{https://plenodb.jpeg.org/pc/upm}.

\bibitem[Bjontegaard(2001)]{bjontegaard2001calculation}
Gisle Bjontegaard.
\newblock Calculation of average psnr differences between rd-curves.
\newblock \emph{VCEG-M33}, 2001.

\bibitem[Cadena et~al.(2016)Cadena, Carlone, Carrillo, Latif, Scaramuzza, Neira, Reid, and Leonard]{cadena2016past}
Cesar Cadena, Luca Carlone, Henry Carrillo, Yasir Latif, Davide Scaramuzza, Jos{\'e} Neira, Ian Reid, and John~J Leonard.
\newblock Past, present, and future of simultaneous localization and mapping: Toward the robust-perception age.
\newblock \emph{IEEE Transactions on robotics}, 32\penalty0 (6):\penalty0 1309--1332, 2016.

\bibitem[Cao et~al.(2019)Cao, Preda, and Zaharia]{cao20193d}
Chao Cao, Marius Preda, and Titus Zaharia.
\newblock 3d point cloud compression: A survey.
\newblock In \emph{The 24th International Conference on 3D Web Technology}, pages 1--9, 2019.

\bibitem[Casado-Coscolla et~al.(2023)Casado-Coscolla, Sanchez-Belenguer, Wolfart, and Sequeira]{casado2023rendering}
Alvaro Casado-Coscolla, Carlos Sanchez-Belenguer, Erik Wolfart, and Vitor Sequeira.
\newblock Rendering massive indoor point clouds in virtual reality.
\newblock \emph{Virtual Reality}, 27\penalty0 (3):\penalty0 1859--1874, 2023.

\bibitem[Chang et~al.(2015)Chang, Funkhouser, Guibas, Hanrahan, Huang, Li, Savarese, Savva, Song, Su, et~al.]{chang2015shapenet}
Angel~X Chang, Thomas Funkhouser, Leonidas Guibas, Pat Hanrahan, Qixing Huang, Zimo Li, Silvio Savarese, Manolis Savva, Shuran Song, Hao Su, et~al.
\newblock Shapenet: An information-rich 3d model repository.
\newblock \emph{arXiv preprint arXiv:1512.03012}, 2015.

\bibitem[Dai et~al.(2017)Dai, Chang, Savva, Halber, Funkhouser, and Nie{\ss}ner]{dai2017scannet}
Angela Dai, Angel~X Chang, Manolis Savva, Maciej Halber, Thomas Funkhouser, and Matthias Nie{\ss}ner.
\newblock Scannet: Richly-annotated 3d reconstructions of indoor scenes.
\newblock In \emph{Proceedings of the IEEE conference on computer vision and pattern recognition}, pages 5828--5839, 2017.

\bibitem[Deitke et~al.(2023)Deitke, Schwenk, Salvador, Weihs, Michel, VanderBilt, Schmidt, Ehsani, Kembhavi, and Farhadi]{deitke2023objaverse}
Matt Deitke, Dustin Schwenk, Jordi Salvador, Luca Weihs, Oscar Michel, Eli VanderBilt, Ludwig Schmidt, Kiana Ehsani, Aniruddha Kembhavi, and Ali Farhadi.
\newblock Objaverse: A universe of annotated 3d objects.
\newblock In \emph{Proceedings of the IEEE/CVF Conference on Computer Vision and Pattern Recognition}, pages 13142--13153, 2023.

\bibitem[d’Eon et~al.(2017)d’Eon, Harrison, Myers, and Chou]{d20178i}
Eugene d’Eon, Bob Harrison, Taos Myers, and Philip~A Chou.
\newblock 8i voxelized full bodies-a voxelized point cloud dataset.
\newblock \emph{ISO/IEC JTC1/SC29 Joint WG11/WG1 (MPEG/JPEG) input document WG11M40059/WG1M74006}, 7\penalty0 (8):\penalty0 11, 2017.

\bibitem[Fan et~al.(2017)Fan, Su, and Guibas]{fan2017point}
Haoqiang Fan, Hao Su, and Leonidas~J Guibas.
\newblock A point set generation network for 3d object reconstruction from a single image.
\newblock In \emph{Proceedings of the IEEE conference on computer vision and pattern recognition}, pages 605--613, 2017.

\bibitem[Galligan et~al.(2018)Galligan, Hemmer, Stava, Zhang, and Brettle]{galligan2018google}
Frank Galligan, Michael Hemmer, Ondrej Stava, Fan Zhang, and Jamieson Brettle.
\newblock Google/draco: a library for compressing and decompressing 3d geometric meshes and point clouds, 2018.

\bibitem[Garrido et~al.(2021)Garrido, Rodrigues, Augusto~Sousa, Jacob, and Castro~Silva]{garrido2021point}
Daniel Garrido, Rui Rodrigues, A Augusto~Sousa, Joao Jacob, and Daniel Castro~Silva.
\newblock Point cloud interaction and manipulation in virtual reality.
\newblock In \emph{2021 5th International Conference on Artificial Intelligence and Virtual Reality (AIVR)}, pages 15--20, 2021.

\bibitem[Geiger et~al.(2013)Geiger, Lenz, Stiller, and Urtasun]{geiger2013vision}
Andreas Geiger, Philip Lenz, Christoph Stiller, and Raquel Urtasun.
\newblock Vision meets robotics: The kitti dataset.
\newblock \emph{The International Journal of Robotics Research}, 32\penalty0 (11):\penalty0 1231--1237, 2013.

\bibitem[Graziosi et~al.(2020)Graziosi, Nakagami, Kuma, Zaghetto, Suzuki, and Tabatabai]{graziosi2020overview}
Danillo Graziosi, Ohji Nakagami, Satoru Kuma, Alexandre Zaghetto, Teruhiko Suzuki, and Ali Tabatabai.
\newblock An overview of ongoing point cloud compression standardization activities: Video-based (v-pcc) and geometry-based (g-pcc).
\newblock \emph{APSIPA Transactions on Signal and Information Processing}, 9:\penalty0 e13, 2020.

\bibitem[He et~al.(2022)He, Ren, Tang, Zhang, Xue, and Fu]{He_2022_CVPR}
Yun He, Xinlin Ren, Danhang Tang, Yinda Zhang, Xiangyang Xue, and Yanwei Fu.
\newblock Density-preserving deep point cloud compression.
\newblock In \emph{Proceedings of the IEEE/CVF Conference on Computer Vision and Pattern Recognition (CVPR)}, 2022.

\bibitem[Ho et~al.(2020)Ho, Jain, and Abbeel]{ho2020denoising}
Jonathan Ho, Ajay Jain, and Pieter Abbeel.
\newblock Denoising diffusion probabilistic models.
\newblock \emph{Advances in neural information processing systems}, 33:\penalty0 6840--6851, 2020.

\bibitem[Hu et~al.(2024)Hu, Ge, Zhao, and Lee]{hu2024x}
Tao Hu, Wenhang Ge, Yuyang Zhao, and Gim~Hee Lee.
\newblock X-ray: A sequential 3d representation for generation.
\newblock \emph{arXiv preprint arXiv:2404.14329}, 2024.

\bibitem[Huang and Liu(2019)]{huang20193d}
Tianxin Huang and Yong Liu.
\newblock 3d point cloud geometry compression on deep learning.
\newblock In \emph{Proceedings of the 27th ACM international conference on multimedia}, pages 890--898, 2019.

\bibitem[Huang et~al.(2022)Huang, Zhang, Chen, Ding, Tai, Zhang, Wang, and Liu]{huang20223qnet}
Tianxin Huang, Jiangning Zhang, Jun Chen, Zhonggan Ding, Ying Tai, Zhenyu Zhang, Chengjie Wang, and Yong Liu.
\newblock 3qnet: 3d point cloud geometry quantization compression network.
\newblock \emph{ACM Transactions on Graphics (TOG)}, 41\penalty0 (6):\penalty0 1--13, 2022.

\bibitem[Jun and Nichol(2023)]{jun2023shap}
Heewoo Jun and Alex Nichol.
\newblock Shap-e: Generating conditional 3d implicit functions.
\newblock \emph{arXiv preprint arXiv:2305.02463}, 2023.

\bibitem[Kerbl et~al.(2023)Kerbl, Kopanas, Leimk{\"u}hler, and Drettakis]{kerbl20233d}
Bernhard Kerbl, Georgios Kopanas, Thomas Leimk{\"u}hler, and George Drettakis.
\newblock 3d gaussian splatting for real-time radiance field rendering.
\newblock \emph{ACM Transactions on Graphics}, 42\penalty0 (4), 2023.

\bibitem[Lester et~al.(2021)Lester, Al-Rfou, and Constant]{lester2021power}
Brian Lester, Rami Al-Rfou, and Noah Constant.
\newblock The power of scale for parameter-efficient prompt tuning.
\newblock \emph{arXiv preprint arXiv:2104.08691}, 2021.

\bibitem[Liu et~al.(2022)Liu, Su, Duanmu, Liu, and Wang]{liu2022perceptual}
Qi Liu, Honglei Su, Zhengfang Duanmu, Wentao Liu, and Zhou Wang.
\newblock Perceptual quality assessment of colored 3d point clouds.
\newblock \emph{IEEE Transactions on Visualization and Computer Graphics}, 29\penalty0 (8):\penalty0 3642--3655, 2022.

\bibitem[Liu et~al.(2023)Liu, Wu, Van~Hoorick, Tokmakov, Zakharov, and Vondrick]{liu2023zero}
Ruoshi Liu, Rundi Wu, Basile Van~Hoorick, Pavel Tokmakov, Sergey Zakharov, and Carl Vondrick.
\newblock Zero-1-to-3: Zero-shot one image to 3d object.
\newblock In \emph{Proceedings of the IEEE/CVF International Conference on Computer Vision}, pages 9298--9309, 2023.

\bibitem[Mao et~al.(2024)Mao, Yuan, Lu, Hamzaoui, and Gao]{mao2024pcac}
Xiaolong Mao, Hui Yuan, Xin Lu, Raouf Hamzaoui, and Wei Gao.
\newblock Pcac-gan: Asparse-tensor-based generative adversarial network for 3d point cloud attribute compression.
\newblock \emph{arXiv preprint arXiv:2407.05677}, 2024.

\bibitem[Michel et~al.(2022)Michel, Bar-On, Liu, Benaim, and Hanocka]{michel2022text2mesh}
Oscar Michel, Roi Bar-On, Richard Liu, Sagie Benaim, and Rana Hanocka.
\newblock Text2mesh: Text-driven neural stylization for meshes.
\newblock In \emph{Proceedings of the IEEE/CVF Conference on Computer Vision and Pattern Recognition}, pages 13492--13502, 2022.

\bibitem[Micikevicius et~al.(2017)Micikevicius, Narang, Alben, Diamos, Elsen, Garcia, Ginsburg, Houston, Kuchaiev, Venkatesh, et~al.]{micikevicius2017mixed}
Paulius Micikevicius, Sharan Narang, Jonah Alben, Gregory Diamos, Erich Elsen, David Garcia, Boris Ginsburg, Michael Houston, Oleksii Kuchaiev, Ganesh Venkatesh, et~al.
\newblock Mixed precision training.
\newblock \emph{arXiv preprint arXiv:1710.03740}, 2017.

\bibitem[Mildenhall et~al.(2020)Mildenhall, Srinivasan, Tancik, Barron, Ramamoorthi, and Ng]{mildenhall2020nerf}
Ben Mildenhall, Pratul~P Srinivasan, Matthew Tancik, Jonathan~T Barron, Ravi Ramamoorthi, and Ren Ng.
\newblock Nerf: Representing scenes as neural radiance fields for view synthesis.
\newblock In \emph{European Conference on Computer Vision}, pages 405--421. Springer, 2020.

\bibitem[Mohammad~Khalid et~al.(2022)Mohammad~Khalid, Xie, Belilovsky, and Popa]{mohammad2022clip}
Nasir Mohammad~Khalid, Tianhao Xie, Eugene Belilovsky, and Tiberiu Popa.
\newblock Clip-mesh: Generating textured meshes from text using pretrained image-text models.
\newblock In \emph{SIGGRAPH Asia 2022 conference papers}, pages 1--8, 2022.

\bibitem[Nguyen and Kaup(2023)]{nguyen2023lossless}
Dat~Thanh Nguyen and Andr{\'e} Kaup.
\newblock Lossless point cloud geometry and attribute compression using a learned conditional probability model.
\newblock \emph{IEEE Transactions on Circuits and Systems for Video Technology}, 33\penalty0 (8):\penalty0 4337--4348, 2023.

\bibitem[Nguyen et~al.(2021)Nguyen, Quach, Valenzise, and Duhamel]{nguyen2021learning}
Dat~Thanh Nguyen, Maurice Quach, Giuseppe Valenzise, and Pierre Duhamel.
\newblock Learning-based lossless compression of 3d point cloud geometry.
\newblock In \emph{ICASSP 2021-2021 IEEE International Conference on Acoustics, Speech and Signal Processing (ICASSP)}, pages 4220--4224. IEEE, 2021.

\bibitem[Nichol et~al.(2022)Nichol, Jun, Dhariwal, Mishkin, and Chen]{nichol2022point}
Alex Nichol, Heewoo Jun, Prafulla Dhariwal, Pamela Mishkin, and Mark Chen.
\newblock Point-e: A system for generating 3d point clouds from complex prompts.
\newblock \emph{arXiv preprint arXiv:2212.08751}, 2022.

\bibitem[Poole et~al.(2022)Poole, Jain, Barron, and Mildenhall]{poole2022dreamfusion}
Ben Poole, Ajay Jain, Jonathan~T Barron, and Ben Mildenhall.
\newblock Dreamfusion: Text-to-3d using 2d diffusion.
\newblock \emph{arXiv preprint arXiv:2209.14988}, 2022.

\bibitem[Qi et~al.(2017{\natexlab{a}})Qi, Su, Mo, and Guibas]{qi2017pointnet}
Charles~R Qi, Hao Su, Kaichun Mo, and Leonidas~J Guibas.
\newblock Pointnet: Deep learning on point sets for 3d classification and segmentation.
\newblock In \emph{Proceedings of the IEEE conference on computer vision and pattern recognition}, pages 652--660, 2017{\natexlab{a}}.

\bibitem[Qi et~al.(2017{\natexlab{b}})Qi, Yi, Su, and Guibas]{qi2017pointnet++}
Charles~Ruizhongtai Qi, Li Yi, Hao Su, and Leonidas~J Guibas.
\newblock Pointnet++: Deep hierarchical feature learning on point sets in a metric space.
\newblock In \emph{Advances in neural information processing systems}, pages 5099--5108, 2017{\natexlab{b}}.

\bibitem[Quach et~al.(2020)Quach, Valenzise, and Dufaux]{quach2020improved}
Maurice Quach, Giuseppe Valenzise, and Frederic Dufaux.
\newblock Improved deep point cloud geometry compression.
\newblock In \emph{2020 IEEE 22nd International Workshop on Multimedia Signal Processing (MMSP)}, pages 1--6. IEEE, 2020.

\bibitem[Rajbhandari et~al.(2020)Rajbhandari, Rasley, Ruwase, and He]{rajbhandari2020zero}
Samyam Rajbhandari, Jeff Rasley, Olatunji Ruwase, and Yuxiong He.
\newblock Zero: Memory optimizations toward training trillion parameter models.
\newblock In \emph{SC20: International Conference for High Performance Computing, Networking, Storage and Analysis}, pages 1--16. IEEE, 2020.

\bibitem[Reddy et~al.(2018)Reddy, Vo, and Narasimhan]{reddy2018carfusion}
N~Dinesh Reddy, Minh Vo, and Srinivasa~G Narasimhan.
\newblock Carfusion: Combining point tracking and part detection for dynamic 3d reconstruction of vehicles.
\newblock In \emph{Proceedings of the IEEE conference on computer vision and pattern recognition}, pages 1906--1915, 2018.

\bibitem[Rombach et~al.(2022)Rombach, Blattmann, Lorenz, Esser, and Ommer]{rombach2022high}
Robin Rombach, Andreas Blattmann, Dominik Lorenz, Patrick Esser, and Bj{\"o}rn Ommer.
\newblock High-resolution image synthesis with latent diffusion models.
\newblock In \emph{Proceedings of the IEEE/CVF conference on computer vision and pattern recognition}, pages 10684--10695, 2022.

\bibitem[Rudolph et~al.(2024{\natexlab{a}})Rudolph, Riemenschneider, and Rizk]{10.1145/3652212.3652217}
Michael Rudolph, Aron Riemenschneider, and Amr Rizk.
\newblock Progressive coding for deep learning based point cloud attribute compression.
\newblock In \emph{Proceedings of the 16th International Workshop on Immersive Mixed and Virtual Environment Systems}, page 78–84, New York, NY, USA, 2024{\natexlab{a}}. Association for Computing Machinery.

\bibitem[Rudolph et~al.(2024{\natexlab{b}})Rudolph, Riemenschneider, and Rizk]{rudolph2024learned}
Michael Rudolph, Aron Riemenschneider, and Amr Rizk.
\newblock Learned compression of point cloud geometry and attributes in a single model through multimodal rate-control.
\newblock \emph{arXiv preprint arXiv:2408.00599}, 2024{\natexlab{b}}.

\bibitem[Ruiz et~al.(2023)Ruiz, Li, Jampani, Pritch, Rubinstein, and Aberman]{ruiz2023dreambooth}
Nataniel Ruiz, Yuanzhen Li, Varun Jampani, Yael Pritch, Michael Rubinstein, and Kfir Aberman.
\newblock Dreambooth: Fine tuning text-to-image diffusion models for subject-driven generation.
\newblock In \emph{Proceedings of the IEEE/CVF conference on computer vision and pattern recognition}, pages 22500--22510, 2023.

\bibitem[Rusu and Cousins(2011)]{rusu20113d}
Radu~Bogdan Rusu and Steve Cousins.
\newblock 3d is here: Point cloud library (pcl).
\newblock In \emph{2011 IEEE international conference on robotics and automation}, pages 1--4. IEEE, 2011.

\bibitem[Saharia et~al.(2022)Saharia, Chan, Saxena, Li, Whang, Denton, Ghasemipour, Gontijo~Lopes, Karagol~Ayan, Salimans, et~al.]{saharia2022photorealistic}
Chitwan Saharia, William Chan, Saurabh Saxena, Lala Li, Jay Whang, Emily~L Denton, Kamyar Ghasemipour, Raphael Gontijo~Lopes, Burcu Karagol~Ayan, Tim Salimans, et~al.
\newblock Photorealistic text-to-image diffusion models with deep language understanding.
\newblock \emph{Advances in Neural Information Processing Systems}, 35:\penalty0 36479--36494, 2022.

\bibitem[Sheng et~al.(2021)Sheng, Li, Liu, Xiong, Li, and Wu]{sheng2021deep}
Xihua Sheng, Li Li, Dong Liu, Zhiwei Xiong, Zhu Li, and Feng Wu.
\newblock Deep-pcac: An end-to-end deep lossy compression framework for point cloud attributes.
\newblock \emph{IEEE Transactions on Multimedia}, 24:\penalty0 2617--2632, 2021.

\bibitem[Singh et~al.(2022)Singh, Singh, Maheshwari, et~al.]{singh2022augmented}
Jaspreet Singh, Gurpreet Singh, Shikha Maheshwari, et~al.
\newblock Augmented reality technology: Current applications, challenges and its future.
\newblock In \emph{2022 4th International Conference on Inventive Research in Computing Applications (ICIRCA)}, pages 1722--1726. IEEE, 2022.

\bibitem[Suzuki et~al.(2022)Suzuki, Karim, Xia, Hedayati, and Marquardt]{suzuki2022augmented}
Ryo Suzuki, Adnan Karim, Tian Xia, Hooman Hedayati, and Nicolai Marquardt.
\newblock Augmented reality and robotics: A survey and taxonomy for ar-enhanced human-robot interaction and robotic interfaces.
\newblock In \emph{Proceedings of the 2022 CHI Conference on Human Factors in Computing Systems}, pages 1--33, 2022.

\bibitem[Tang et~al.(2023)Tang, Ren, Zhou, Liu, and Zeng]{tang2023dreamgaussian}
Jiaxiang Tang, Jiawei Ren, Hang Zhou, Ziwei Liu, and Gang Zeng.
\newblock Dreamgaussian: Generative gaussian splatting for efficient 3d content creation.
\newblock \emph{arXiv preprint arXiv:2309.16653}, 2023.

\bibitem[Tian et~al.(2017)Tian, Ochimizu, Feng, Cohen, and Vetro]{tian2017geometric}
Dong Tian, Hideaki Ochimizu, Chen Feng, Robert Cohen, and Anthony Vetro.
\newblock Geometric distortion metrics for point cloud compression.
\newblock In \emph{2017 IEEE International Conference on Image Processing (ICIP)}, pages 3460--3464. IEEE, 2017.

\bibitem[Tochilkin et~al.(2024)Tochilkin, Pankratz, Liu, Huang, Letts, Li, Liang, Laforte, Jampani, and Cao]{tochilkin2024triposr}
Dmitry Tochilkin, David Pankratz, Zexiang Liu, Zixuan Huang, Adam Letts, Yangguang Li, Ding Liang, Christian Laforte, Varun Jampani, and Yan-Pei Cao.
\newblock Triposr: Fast 3d object reconstruction from a single image.
\newblock \emph{arXiv preprint arXiv:2403.02151}, 2024.

\bibitem[Wang et~al.(2023)Wang, Du, Li, Yeh, and Shakhnarovich]{wang2023score}
Haochen Wang, Xiaodan Du, Jiahao Li, Raymond~A Yeh, and Greg Shakhnarovich.
\newblock Score jacobian chaining: Lifting pretrained 2d diffusion models for 3d generation.
\newblock In \emph{Proceedings of the IEEE/CVF Conference on Computer Vision and Pattern Recognition}, pages 12619--12629, 2023.

\bibitem[Wang and Ma(2022)]{wang2022sparsea}
Jianqiang Wang and Zhan Ma.
\newblock Sparse tensor-based point cloud attribute compression.
\newblock In \emph{2022 IEEE 5th International Conference on Multimedia Information Processing and Retrieval (MIPR)}, pages 59--64. IEEE, 2022.

\bibitem[Wang et~al.(2021{\natexlab{a}})Wang, Ding, Li, and Ma]{wang2021multiscale}
Jianqiang Wang, Dandan Ding, Zhu Li, and Zhan Ma.
\newblock Multiscale point cloud geometry compression.
\newblock In \emph{2021 Data Compression Conference (DCC)}, pages 73--82. IEEE, 2021{\natexlab{a}}.

\bibitem[Wang et~al.(2021{\natexlab{b}})Wang, Zhu, Liu, and Ma]{wang2021lossy}
Jianqiang Wang, Hao Zhu, Haojie Liu, and Zhan Ma.
\newblock Lossy point cloud geometry compression via end-to-end learning.
\newblock \emph{IEEE Transactions on Circuits and Systems for Video Technology}, 31\penalty0 (12):\penalty0 4909--4923, 2021{\natexlab{b}}.

\bibitem[Wang et~al.(2022)Wang, Ding, Li, Feng, Cao, and Ma]{wang2022sparse}
Jianqiang Wang, Dandan Ding, Zhu Li, Xiaoxing Feng, Chuntong Cao, and Zhan Ma.
\newblock Sparse tensor-based multiscale representation for point cloud geometry compression.
\newblock \emph{IEEE Transactions on Pattern Analysis and Machine Intelligence}, 45\penalty0 (7):\penalty0 9055--9071, 2022.

\bibitem[Wiesmann et~al.(2021)Wiesmann, Milioto, Chen, Stachniss, and Behley]{wiesmann2021deep}
Louis Wiesmann, Andres Milioto, Xieyuanli Chen, Cyrill Stachniss, and Jens Behley.
\newblock Deep compression for dense point cloud maps.
\newblock \emph{IEEE Robotics and Automation Letters}, 6\penalty0 (2):\penalty0 2060--2067, 2021.

\end{thebibliography}
}

\clearpage
\setcounter{page}{1}

\maketitlesupplementary

    \begin{figure*}[htbp]
        \centering
        \includegraphics[width=1.0\textwidth, keepaspectratio]{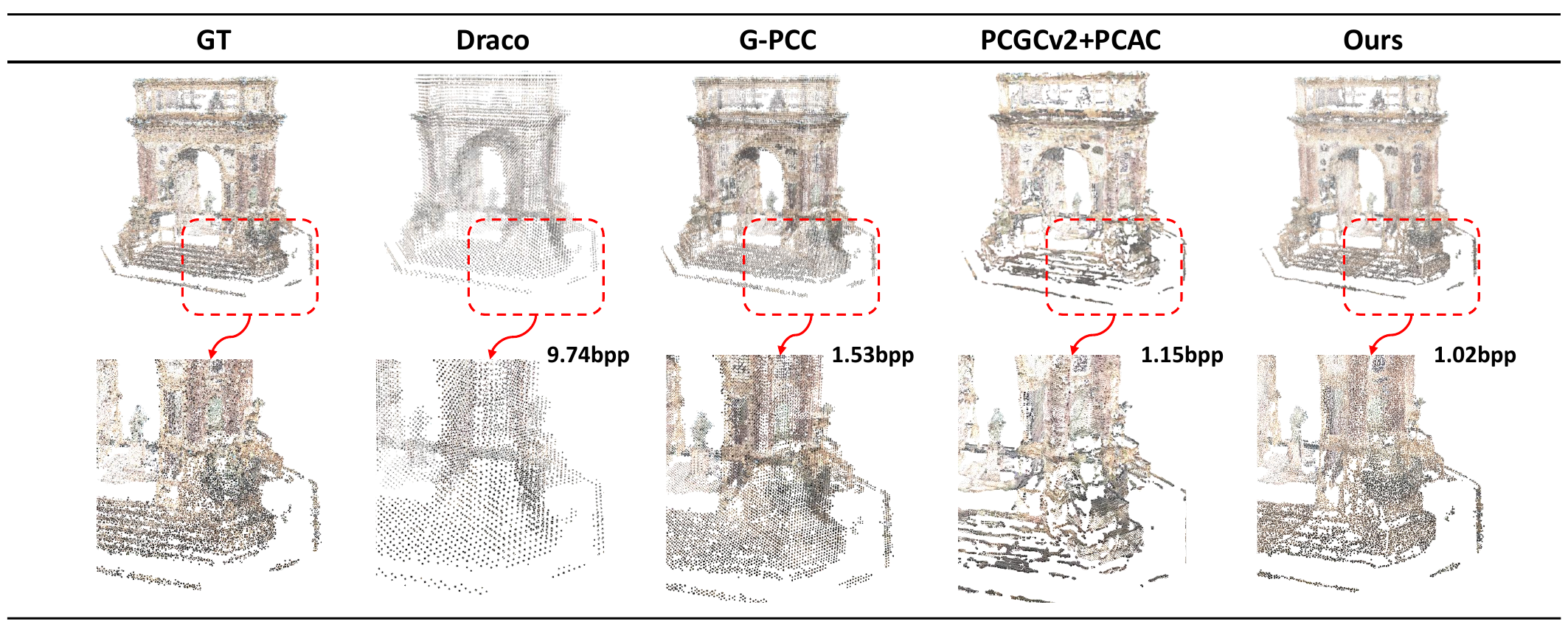}
        \caption{Qualitative comparisons on Noisy Real Scans.}
        \label{pic:buildings}
    \end{figure*}

\section{Details about the Settings}
\label{sec:settings}
Table~\ref{tab:para} provides detailed information on all hyperparameters discussed in Sec.~\ref{sec:method}. Our experiments are conducted on RTX 4090ti/A5000 GPUs using PyTorch 2.0.0 and CUDA 11.6. For the KNN operation mentioned in the Weights Aggregation module in Sec.~\ref{sec:method}, we adopt the ball query algorithm proposed in \cite{qi2017pointnet++}. This algorithm selects \( k \) neighbors within a ball of radius \( r \), ensuring the KNN operation is constrained to a local scale.

	\begin{table}[ht]
 \vskip -0.05in
		\normalsize
		\begin{center}
			\scalebox{1.0}{
				\begin{tabular}{cc}
\toprule
            & Settings                                \\ \cmidrule{2-2}
Optimization       & Adam Optimizer, lr 5e-5  \\
Number of $k$, radius $r$    & 32, 0.004                                \\
Point-E~\cite{nichol2022point} model & Base40M        \\
Iterations & 1000                                     \\ 
$T$ & 8                                                \\
\bottomrule
\end{tabular}
			}
		\end{center}
        \vskip -0.2in
        \caption{The detailed setting of mentioned hyper-parameters.}
        \label{tab:para}
        \vskip -0.25in
	\end{table}

\section{Comparison on Denser Point Clouds}
To further validate the performances of our method, we also conduct experiments on denser point clouds from GIT-UPM dataset~\cite{buildings} and 8iVFB~\cite{d20178i}. GIT-UPM dataset~\cite{buildings} includes large scale outdoor noisy point clouds, while 8iVFB composes of voxelized full human bodies. For the convience of comparison, we sample point clouds from both dataset uniformly to 640k points for comparison.

\textbf{Comparisons on Noisy Real Scans}
Qualitative and quantitative comparisons for the GIT-UPM dataset~\cite{buildings} are shown in Fig.~\ref{pic:buildings} and Table~\ref{tab:buildings}, respectively. 
We can observe that our method still outperforms the baselines in the geometric metrics $D_1$ and $D_2$, while the performances in the color metric decrease slightly. 
\begin{table}[ht]
        \centering
        \normalsize

        \setlength\tabcolsep{1.5pt}
  \scalebox{0.7}{
\begin{tabular}{c|ccccc}
\toprule
Objects                       &    & PCGCv2+PCAC     & G-PCC           & Draco & SparsePCGC+PCAC    \\ \midrule
\multirow{3}{*}{Arco}&D1&+2.19(12.25\%)&+2.76(14.21\%)&+28.59(55.37\%)&+5.61(28.29\%) \\ 
&D2&+0.79(3.71\%)&+1.81(7.92\%)&+27.68(54.37\%)&+7.20(30.65\%) \\ 
&D3&-1.17(5.61\%)&-1.99(9.18\%)&+6.27(26.40\%)&-0.18(0.84\%)\\ \midrule
\multirow{3}{*}{Villa}&D1&-1.58(7.40\%)&+3.65(16.79\%)&+25.45(55.29\%)&+10.75(47.86\%) \\ 
&D2&-1.83(7.28\%)&+2.29(9.23\%)&+24.52(57.60\%)&+15.76(59.79\%) \\ 
&D3&-1.45(7.59\%)&-1.38(7.11\%)&+3.33(17.30\%)&+1.16(6.04\%)\\ \midrule
\multirow{3}{*}{Palazzo}&D1&+3.96(11.47\%)&+3.91(11.05\%)&+23.35(54.18\%)&-6.83(19.13\%) \\ 
&D2&+3.09(8.20\%)&+2.57(6.68\%)&+21.99(58.51\%)&-8.53(21.80\%) \\ 
&D3&-0.96(5.14\%)&-1.81(9.58\%)&+3.05(16.61\%)&-0.99(5.21\%)\\ \midrule
\multirow{3}{*}{Aver}&D1&+1.52(5.44\%)&+3.44(14.02\%)&+25.80(54.95\%)&+3.18(19.01\%) \\ 
&D2&+0.68(1.54\%)&+2.22(7.94\%)&+24.73(56.82\%)&+4.81(22.88\%) \\ 
&D3&-1.19(6.11\%)&-1.73(8.63\%)&+4.22(20.10\%)&+0.00(0.00\%)\\ \bottomrule
\end{tabular}
}
\vskip -0.05in
\caption{Quantitative comparisons on outdoor buildings.  Items in the table denote the relative improvements of our method over other methods measured with BD-PSNR. '+' and '-' indicates performance increases and decreases, respectively.}
\label{tab:buildings}
\end{table}
\begin{table}[ht]
        \centering
        \normalsize

        \setlength\tabcolsep{1.5pt}
  \scalebox{0.65}{
\begin{tabular}{c|ccccc}
\toprule
Objects                       &    & PCGCv2+PCAC     & G-PCC           & Draco & SparsePCGC+PCAC    \\ \midrule
\multirow{3}{*}{soldier}&D1&+2.42(18.87\%)&+5.40(39.99\%)&+36.96(63.11\%)&+0.88(7.39\%) \\ 
&D2&+0.87(5.87\%)&+3.95(24.33\%)&+32.81(60.81\%)&-0.65(4.82\%) \\ 
&D3&-2.47(7.78\%)&-4.65(14.15\%)&+15.19(40.97\%)&-4.18(13.43\%)\\ \midrule
\multirow{3}{*}{redandblack}&D1&-0.22(1.94\%)&+5.89(48.37\%)&+28.50(58.03\%)&-2.94(29.18\%) \\ 
&D2&-1.65(13.82\%)&+3.73(26.03\%)&+25.08(57.20\%)&-5.61(50.86\%) \\ 
&D3&-4.56(16.48\%)&-5.31(17.61\%)&+8.22(29.41\%)&-5.32(19.36\%)\\ \midrule
\multirow{3}{*}{longdress}&D1&+1.09(7.70\%)&+6.84(46.24\%)&+30.00(60.75\%)&+0.35(2.62\%) \\ 
&D2&+0.87(5.41\%)&+5.91(33.27\%)&+27.09(62.66\%)&+0.67(4.39\%) \\ 
&D3&-2.91(11.01\%)&-1.14(4.06\%)&+8.05(31.12\%)&-3.50(13.48\%)\\ \midrule
\multirow{3}{*}{loot}&D1&+0.61(5.04\%)&+6.97(54.00\%)&+29.67(58.46\%)&+2.36(21.47\%) \\ 
&D2&+0.20(1.43\%)&+5.94(37.45\%)&+26.06(58.26\%)&+1.77(14.20\%) \\ 
&D3&-3.04(8.60\%)&-2.77(7.64\%)&+10.81(30.99\%)&-4.32(12.57\%)\\ \midrule
\multirow{3}{*}{Aver}&D1&+0.97(7.42\%)&+6.27(47.15\%)&+31.28(60.09\%)&+0.16(0.57\%) \\ 
&D2&+0.08(0.28\%)&+4.88(30.27\%)&+27.76(59.73\%)&-0.96(9.27\%) \\ 
&D3&-3.24(10.96\%)&-3.47(10.87\%)&+10.57(33.12\%)&-4.33(14.71\%)\\ \bottomrule
\end{tabular}
}
\vskip -0.05in
\caption{Quantitative comparisons on 8iVFB.  Items in the table denote the relative improvements of our method over other methods measured with BD-PSNR. '+' and '-' indicates performance increases and decreases, respectively.}
\vskip -0.1in
\label{tab:voxel}
\end{table}
The decrease in color compression performance may be attributed to color discontinuities on surfaces, particularly in large-scale point clouds where colors vary significantly across regions. The denoising process in the diffusion model tends to predict a continuous color distribution, which can slightly deviate from the original noisy colors in outdoor data. This makes our color compression less effective in regions with complex color variations. In contrast, encoding colors separately as supplementary features, as done in PCGCv2~\cite{wang2021multiscale}, may be more suitable for such cases.
As shown in Fig.~\ref{pic:buildings}, our method reconstructs clearer structures at a lower bpp, demonstrating its effectiveness for dense and noisy point clouds.

\textbf{Comparison on Voxelized Points}
Another widely used benchmark for point cloud compression is the voxelized 8iVFB~\cite{d20178i}, which consists of multiple human body models represented by 3D voxelized points rather than discrete points, as in previous experiments. In this section, we present a quantitative comparison with existing baselines in Table~\ref{tab:voxel}. While our method underperforms in color metrics compared to the baselines, it achieves the best results in most geometric metrics.
This discrepancy may be attributed to the characteristics of 8iVFB~\cite{d20178i}. First, the models in 8iVFB exhibit abrupt color changes, making them challenging for diffusion-based approaches to model effectively. Second, 
SparsePCGC~\cite{wang2022sparse} and PCAC~\cite{mao2024pcac} are specifically trained on voxelized data, whereas Point-E~\cite{nichol2022point} has not been trained or fine-tuned on such point clouds, leading to its comparatively lower performance.

Nevertheless, our method’s superior performance on geometric metrics demonstrates its robustness in handling voxelized point clouds. Former extensive experiments on objects, indoor scenes, and outdoor buildings have also confirmed its effectiveness on point clouds with diverse distributions.
The color compression performance could be further improved by incorporating additional color features using methods like PCAC~\cite{mao2024pcac}, and the overall performances on voxelized data may be futher improved by introducing some test-time fine-tuning to Point-E~\cite{nichol2022point} on single data with methods like Dreambooth~\cite{ruiz2023dreambooth}. We will explore these approaches in our future work.

\end{document}